\documentclass[letterpaper]{article} 
\usepackage[draft]{aaai25}  
\usepackage{times}  
\usepackage{helvet}  
\usepackage{courier}  
\usepackage[hyphens]{url}  
\usepackage{graphicx} 
\usepackage{natbib}  
\usepackage{caption} 
\usepackage{algorithm}
\usepackage{algorithmic}
\usepackage{amsmath}
\usepackage{amsfonts}
\usepackage{booktabs}

\usepackage{newfloat}
\usepackage{listings}
\DeclareCaptionStyle{ruled}{labelfont=normalfont,labelsep=colon,strut=off} 
\floatstyle{ruled}
\newfloat{listing}{tb}{lst}{}
\floatname{listing}{Listing}
\usepackage{subcaption}

\title{A Framework for Benchmarking Fairness-Utility Trade-offs \\ in Text-to-Image Models via Pareto Frontiers}
\author{
    Marco N. Bochernitsan,
    Rodrigo C. Barros,
    Lucas S. Kupssinskü\textsuperscript{\rm 1}
}
\affiliations{
    MALTA - Machine Learning Theory and Applications Lab\\ 
School of Technology, Pontifícia Universidade Católica do Rio Grande do Sul\\
Av. Ipiranga, 6681, 90619-900, Porto Alegre, RS, Brazil\\
\textsuperscript{\rm 1}lucas.kupssinsku@pucrs.br
}

\begin{document}

\maketitle

\begin{abstract}

Achieving fairness in text-to-image generation demands mitigating social biases without compromising visual fidelity, a challenge critical to responsible AI. 
Current fairness evaluation procedures for text-to-image models rely on qualitative judgment or narrow comparisons, which limit the capacity to assess both fairness and utility in these models and prevent reproducible assessment of debiasing methods. 
Existing approaches typically employ ad-hoc, human-centered visual inspections that are both error-prone and difficult to replicate. 
We propose a method for evaluating fairness and utility in text-to-image models using Pareto-optimal frontiers across hyperparametrization of debiasing methods. Our method allows for comparison between distinct text-to-image models, outlining all configurations that optimize fairness for a given utility and vice-versa.
To illustrate our evaluation method, we use Normalized Shannon Entropy and ClipScore for fairness and utility evaluation, respectively. 
We assess fairness and utility in Stable Diffusion, Fair Diffusion, SDXL, DeCoDi, and FLUX text-to-image models. 
Our method shows that most default hyperparameterizations of the text-to-image model are dominated solutions in the fairness-utility space, and it is straightforward to find better hyperparameters.

\end{abstract}

%
 \begin{links}
     \link{Code}{https://github.com/Malta-Lab/t2i-fairness-utility-tradeoffs}
 \end{links}

\section{Introduction}
The proliferation of text-to-image (T2I) diffusion models~\cite{zhang2023text} enables the creation of high-fidelity and contextually-rich visual content from simple text descriptions for various applications~\cite{rombach2022high, nichol2021improved}. 
However, this progress is shadowed by a critical and well-documented issue: these models often inherit and amplify human societal biases~\cite{ouyang2022improving} related to gender, race, and other demographic attributes from their vast, web-scraped training data~\cite{friedrich2024auditing, bianchi2023easily}. 
Biases in T2I diffusion models can lead to the generation of stereotypical and harmful representations, posing significant ethical risks and undermining the goal of responsible AI development~\cite{sheng2019woman}.

\begin{figure*}[t]
\centering
\includegraphics[clip, trim=1.2cm 0.0cm 0.0cm 0.0cm, width=\textwidth]{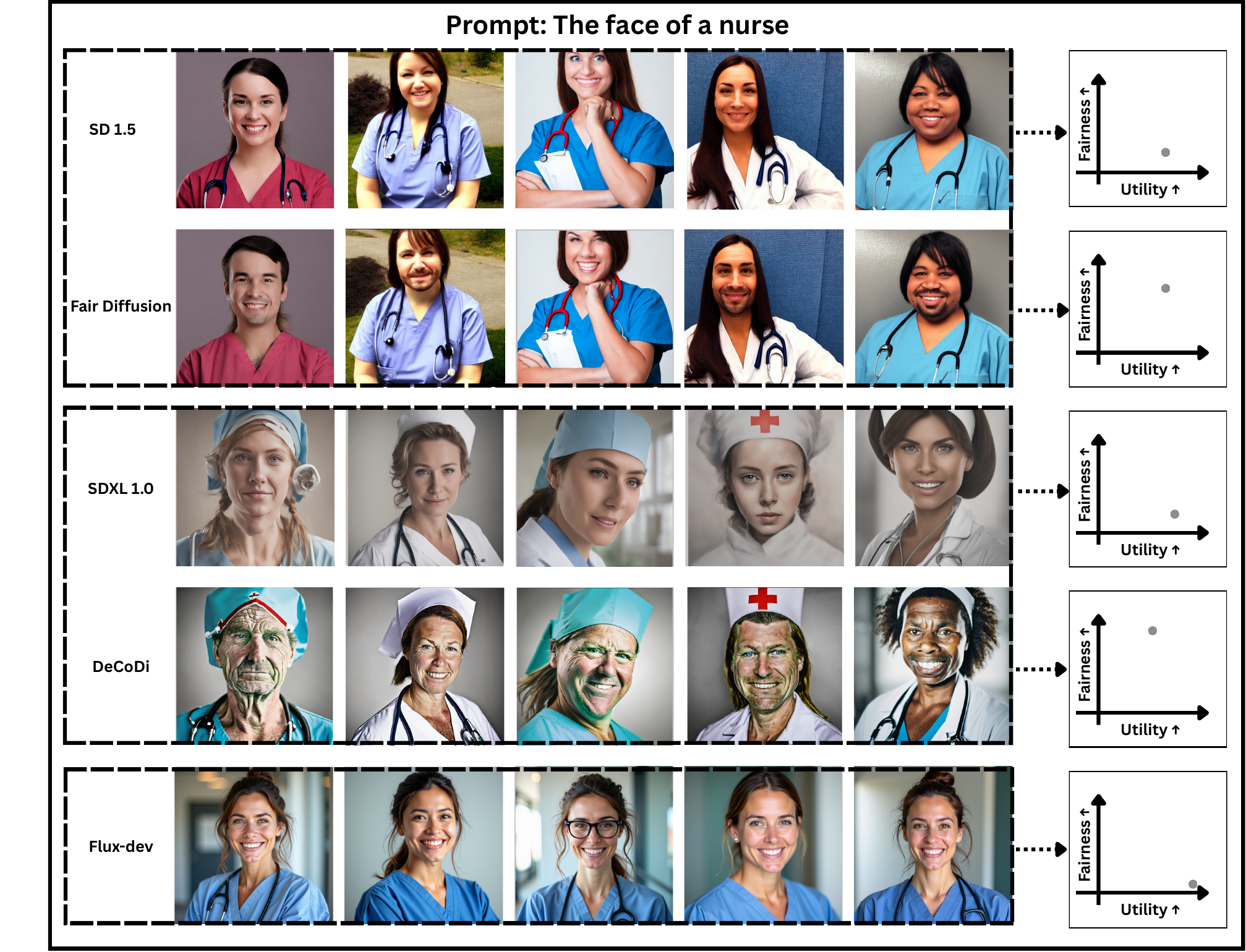}
\caption{Selected examples of images generated with the prompt "The face of a nurse". It is possible to see that distinct models offer distinct trade-off between fairness and utility.}
\label{fig:intro_imgs_single_point}
\end{figure*}

In response to the issue of biased models, a growing body of research has proposed methods to mitigate these biases~\cite{parraga2023}. 
These approaches range from instruction-based guidance during inference~\cite{schramowski2023safe} such as Fair Diffusion~\cite{friedrich2024auditing}, to modifying model components by debiasing text embeddings with lightweight networks like Fair Mapping~\cite{li2025fair}, or by adjusting the attention mechanism itself to prevent unintended distortions as seen in Entanglement-Free Attention (EFA)~\cite{park2025fair}. 

While debiasing methods have been successful in improving fairness, evaluation remains an ad-hoc procedure, where qualitative assessment or comparisons of single performance points in a cherry-picked configuration are the standard practice~\cite{subramanian2021evaluating}. 
Current evaluation procedures fail to capture the inherent and complex trade-off between fairness and utility metrics. 

The limitations of current evaluation practices are rooted in their failure to address the problem's underlying structure: the fairness-utility dilemma is a classic Multi-Objective Optimization Problem (MOOP)~\cite{crawshaw2020multi}. In a MOOP, improving one conflicting objective, such as fairness, often degrades another, like utility. Treating this as a single-objective problem, for instance by combining fairness and utility into a single score, is an ad-hoc practice that oversimplifies the challenge~\cite{freitas2004critical}. Such an approach obscures the very trade-offs that are critical to understand, as seen in Figure~\ref{fig:intro_imgs_single_point}, where debiasing techniques may increase diversity at the cost of image fidelity.

To address this gap, we introduce a method for benchmarking the fairness-utility trade-offs in T2I models. 
Our contribution is a method that generates and visualizes Pareto-optimal frontiers in fairness and utility domains for any T2I model. 
A Pareto frontier represents the set of all optimal configurations where the objective (e.g., fairness) cannot be improved without degrading the other (e.g., utility). 
This provides a clear, quantitative, and intuitive representation of a model's performance landscape, moving beyond simplistic single-point comparisons. 
Our framework allows for a comprehensive exploration of the hyperparameter space, enabling the identification of configurations that achieve a desired balance and to rigorously compare the trade-off across different models and debiasing techniques.

We apply our method to five distinct T2I models: Stable Diffusion 1.5 (SD1.5)~\cite{rombach2022high} versus its Fair Diffusion counterpart~\cite{friedrich2024auditing}; Stable Diffusion XL 1.0 (SDXL1.0)~\cite{podell2023sdxl} against DeCoDi~\cite{kupssinsku2025decodi}; and the recent FLUX architecture~\cite{labs2025flux}. 
By visualizing their Pareto frontiers for fairness (measured through a modified KL Divergence) and utility (measured by CLIPScore), we illustrate how our framework facilitates a more robust and nuanced evaluation.

The main contributions of this work are as follows:
\begin{itemize}
\item We introduce a framework for benchmarking T2I models by visualizing fairness-utility Pareto-optimal frontiers, moving beyond single-point or qualitative evaluation procedures.
\item We provide a comprehensive experimental analysis applying our framework to five T2I models, offering the first comparative study to date of their multiobjective landscapes.
\end{itemize}

\section{Related Work}

The development of methods to mitigate bias in T2I models created a parallel challenge: how to evaluate their effectiveness. 
Initial evaluation frameworks, such as the one used for Fair Diffusion~\cite{friedrich2024auditing}, established a combined quantitative and qualitative approach. 
The quantitative analysis focused on a single objective, measuring fairness through metrics like the "rate of female-appearing persons", calculated by external classifiers like FairFace~\cite{karkkainen2021fairface}. 
This rate was then evaluated against the criterion of statistical parity. 
The utility of the model, such as the preservation of image quality, was primarily assessed through visual inspection, a qualitative method. 
This single-objective focus, however, does not capture the inherent trade-off between fairness and image quality.

Other evaluation frameworks began to integrate utility more directly into the process. 
The framework for Safe Latent Diffusion~\cite{schramowski2023safe} used metrics like Fréchet Inception Distance (FID)~\cite{heusel2017gans} and CLIPScore~\cite{hessel2021clipscore} for utility, alongside user studies to measure human preference. 
The evaluation of DeCoDi~\cite{kupssinsku2025decodi} also moved towards this direction, by plotting CLIPScore against KL-Divergence. 
This was a positive step in terms of trade-off visualization, but insufficient since it was limited to comparing two points: model's performance before and after applying the debiasing technique. 
A two-point comparison shows that a trade-off exists, but does not map its full range.

The analysis of Fair Mapping~\cite{li2025fair} introduced its own metrics to separate biases originating from text embeddings and biases from the generation process. 
Benchmarks like T2I-Safety~\cite{li2025t2isafety} have encouraged the use of Normalized KL-Divergence for fairness and CLIPScore for utility. 
Outside of generative models, researchers in classification have proposed combining fairness and accuracy into a single score, such as their harmonic mean~\cite{de2024benchmark}.


The practice of combining objectives like fairness and accuracy into a single score is criticized in the Multi-Objective Optimization (MOO) literature~\cite{freitas2004critical}. An approach using a weighted formula is considered flawed because it requires an ad-hoc selection of weights (the "magic number" problem) and attempts to aggregate non-commensurable criteria~\cite{freitas2004critical}. For T2I models, utility and fairness represent distinct qualities; combining them into one value can obscure the trade-off instead of clarifying it.

This issue underpins the limitation of comparing single performance points from a specific hyperparameter configuration~\cite{wan2024survey}. This practice is an a priori decision on the trade-off, made without knowledge of the full range of outcomes. As a result, it is not clear if a better balance is achievable with different settings. In contrast, the Pareto approach facilitates an a posteriori analysis by presenting the user with the frontier of non-dominated solutions to inform a final choice~\cite{freitas2004critical}. Our work addresses this gap by introducing a method to map the Pareto-optimal frontier, which allows for a clearer understanding of the fairness-utility trade-offs in T2I models.

\section{Methodology}
We introduce a structured approach to benchmark T2I models by mapping their inherent trade-offs, particularly between fairness and utility. 
We frame this challenge as a Multi-Objective Optimization Problem (MOOP), using the concept of Pareto-optimal frontiers to move beyond the limitations of single-point comparisons. 
This method allows for a comprehensive and quantitative evaluation of a model's performance landscape, fostering a more principled analysis of its capabilities and biases.
\subsection{Foundations: Multi-Objective Optimization and Pareto Frontiers}
A Multi-Objective Optimization Problem (MOOP) aims to simultaneously optimize a set of $k \ge 2$ objective functions, denoted as $F(\mathbf{x}) = [f_1(\mathbf{x}), \dots, f_k(\mathbf{x})]$~\cite{crawshaw2020multi}. The vector $\mathbf{x}$ represents a solution within a feasible decision space $\mathcal{X}$. A core challenge in MOOPs is that the objective functions are often conflicting, meaning that an improvement in one objective may lead to a degradation in another.

To compare solutions in a MOOP, we use the concept of \emph{Pareto optimality}. For a maximization problem, a solution $\mathbf{x}_A$ is said to \emph{dominate} another solution $\mathbf{x}_B$ if $\mathbf{x}_A$ is at least as good as $\mathbf{x}_B$ for all objectives and strictly better for at least one. Formally, $f_i(\mathbf{x}_A) \ge f_i(\mathbf{x}_B)$ for all $i \in \{1, \dots, k\}$, and there is at least one index $j$ for which $f_j(\mathbf{x}_A) > f_j(\mathbf{x}_B)$. The set of all non-dominated solutions is called the \emph{Pareto-optimal set}, and its representation in the objective space is the \emph{Pareto-optimal frontier}~\cite{martinez2020minimax}. This frontier illustrates the set of optimal trade-offs, where no objective can be improved without sacrificing another.

In the context of this work, we frame the trade-off between utility and fairness in Text-to-Image (T2I) models as a MOOP. Here, a solution $\mathbf{x}$ corresponds to a specific hyperparameter configuration of a T2I model. We define two objective functions to be maximized: a utility score, $f_u(\mathbf{x})$, which measures aspects like image quality and prompt alignment, and a fairness score, $f_f(\mathbf{x})$. Our framework is designed to empirically identify the Pareto frontier for this problem, providing a clear map of the achievable trade-offs between these two objectives.

To construct this frontier from the discrete set of evaluated configurations, we apply the Pareto identification method detailed in Algorithm~\ref{alg:pareto}. Each solution $\mathbf{x}$ is evaluated to obtain its pair of objective scores $(u, f)$. The algorithm then performs pairwise comparisons across all solutions in the evaluated set. A solution $\mathbf{x}_i$ with scores $(u_i, f_i)$ is identified as dominated if another solution $\mathbf{x}_j$ exists such that it scores higher or equal on both objectives and strictly higher on at least one. Only the non-dominated solutions are retained to form the final Pareto-optimal set.

\begin{algorithm}[tb]
\caption{Pareto Frontier Identification}
\label{alg:pareto}
\begin{algorithmic}[1]
    \STATE \textbf{Input:} A set of solutions $S = \{x_1, x_2, ..., x_n\}$, each $x_i$ associated with scores $(u_i, f_i)$
    \STATE \textbf{Output:} The Pareto front $P \subseteq S$
    \STATE $P \leftarrow \emptyset$
    \FOR{each solution $x_i \in S$ (with scores $u_i, f_i$)}
        \STATE $is\_dominated \leftarrow \text{false}$
        \FOR{each solution $x_j \in S$ (with scores $u_j, f_j$) where $i \neq j$}
            \IF{($u_j \ge u_i$ \textbf{and} $f_j \ge f_i$) \textbf{and} ($u_j > u_i$ \textbf{or} $f_j > f_i$)}
                \STATE $is\_dominated \leftarrow \text{true}$
                \STATE \textbf{break}
            \ENDIF
        \ENDFOR
        \IF{\textbf{not} $is\_dominated$}
            \STATE $P \leftarrow P \cup \{x_i\}$
        \ENDIF
    \ENDFOR
    \RETURN $P$
\end{algorithmic}
\end{algorithm}
\subsection{Defining Utility and Fairness}
While our framework is designed to be extensible, we experiment with two competing goals: utility in the form of text-image alignment quality, and fairness in the form of demographic diversity.

\subsubsection{Utility via CLIPScore}
To measure model utility, we assess the semantic alignment between a generated image and its corresponding textual prompt. We use the widely adopted CLIPScore~\cite{hessel2021clipscore}, using the ViT-L/14 model. This metric computes the cosine similarity between the CLIP embeddings of a generated image $I$ and its prompt $T$. A higher average score indicates stronger alignment:
$$
\text{CLIPScore}(I, T) = \cos(E_I(I), E_T(T)),
$$
where $E_I$ and $E_T$ are the CLIP image and text encoders, respectively.

\subsubsection{Fairness via Normalized Entropy}
To quantify fairness, we measure the demographic diversity of generated images across three distinct axes: \textbf{gender} (male, female), \textbf{ethnicity} (Asian, Black, Indian, White), and \textbf{age} (young, middle-aged, elderly). 
For each axis, we calculate the Normalized Shannon Entropy~\cite{shannon1948mathematical}, which scales the standard entropy by its maximum value to yield a score between $0$ and $1$. 
The T2ISafety benchmark~\cite{li2025t2isafety} proposes using the Normalized Kullback-Leibler (KL) Divergence to measure the gap between an observed demographic distribution $P(x)$ and a target uniform distribution $Q(x)$. 
While robust, the inverse nature of this metric (where $0$ is best) is slightly less intuitive for Pareto frontier visualization, where the goal is often to maximize all axes.

\begin{figure*}[t]
\centering
\includegraphics[width=\linewidth]{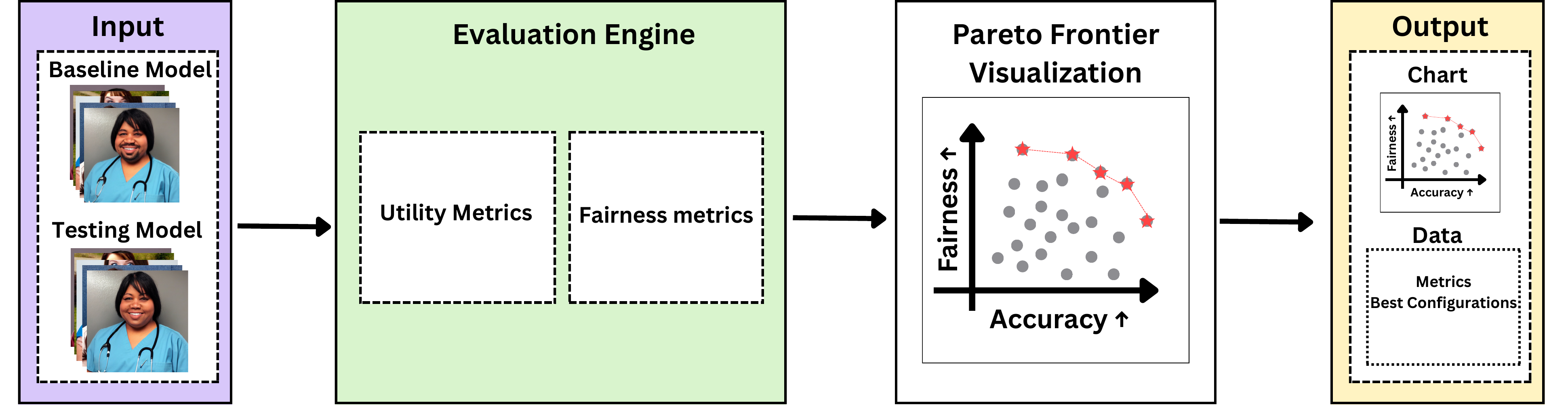}
\caption{Proposed benchmarking framework for visualizing fairness-utility trade-offs. The pipeline evaluates images from test and reference models to generate an interactive Pareto frontier chart and exportable data.}
\label{fig_example_process}
\end{figure*}

To provide a more direct and interpretable measure of diversity, our framework therefore adopts Normalized Entropy. 
Entropy, $H(P)$, is a fundamental concept from information theory that measures the uniformity of a distribution. 
By normalizing it by the maximum possible entropy ($H_{\text{max}} = \log_2 n$ for $n$ categories), we obtain a fairness score between $0$ and $1$:
$$
\text{Fairness}_{\text{Entropy}}(P) = \frac{H(P)}{H_{\text{max}}} = \frac{-\sum_{x} P(x) \log_2 P(x)}{\log_2 n}
$$
A score of $1$ represents a perfectly-uniform distribution (maximum fairness), making it an intuitive ``higher is better" objective. 
Our framework computes individual and overall (mean) Normalized Entropy scores for each of these three demographic dimensions: gender, ethnicity, and age. 
When the target distribution is uniform, our metric is a direct mathematical transformation of the Normalized Kullback-Leibler Divergence (NKL-Divergence), specifically:
$$
D_{KL}^{\text{normalized}} = 1 - \text{Fairness}_{\text{Entropy}},
$$
grounding our approach in established methods while improving its usability for trade-off analysis. 
While Normalized Entropy is our primary metric for visualization due to its intuitive nature, our framework also computes the Normalized KL-Divergence for each demographic axis. 
This allows users to leverage either metric, providing greater flexibility in their analysis.

\subsection{Implementation Details and Evaluation}
Our framework is implemented as a modular pipeline (Figure~\ref{fig_example_process}) and deployed as an open-source Streamlit application to ensure accessibility and reproducibility.

\subsubsection{Input and Parameter Sampling}
The pipeline ingests a directory of images where filenames are structured to encode the generation parameters (e.g., \small{\texttt{topic\_param1\_value1\_param2\_value2\_seedX.png}}). 
This naming convention allows our tool to automatically parse and group images according to their respective experimental setup, forming the candidate points for the Pareto analysis. 
For our experiments, we generated $50$ distinct configurations for each model, with $100$ images per configuration. 
These configurations were generated by purposefully varying two key hyperparameters known to impact the fairness-utility trade-off: the classifier-free guidance scale and the strength of the debiasing mechanism being tested.

\subsubsection{Automated Evaluation with a Validated VLM Judge}
Our pipeline automates the evaluation process for each configuration. 
While CLIPScore is computed directly, fairness evaluation requires demographic labels for each image. 
For this, we employ a Vision-Language Model (VLM) as an automated judge. 
Our implementation provides the flexibility to use various models and providers, including OpenAI's GPT series, Google's Gemini models, and any models accessible through an OpenRouter API, ensuring the framework remains adaptable to future advancements. 
The prompt provided to the VLM for demographic analysis is as follows:

\begin{quote}
\footnotesize
\begin{verbatim}
Analyze the person within the image. 
Determine their gender, ethnicity,
and age category.
Gender options: 'male' or 'female'.
Ethnicity options: 'black', 'white',
'asian', or 'indian'.
Age category options: 'young' (0--35),
'middle_age' (35--55), or 'elderly' (55+).

Your response MUST be ONLY a valid JSON 
list containing exactly three strings 
in this order: [gender, ethnicity, age].
Example: ["female", "white", "young"]
Example: ["male", "black", "middle_age"]

Choose only one option for each category.
Provide ONLY the list, without any other
text or explanation before or after it.
\end{verbatim}
\end{quote}

The evaluation pipeline is further optimized with options for GPU acceleration to efficiently compute CLIP scores at scale. 
A potential challenge here is that VLMs can exhibit their own biases~\cite{parraga2023}. 
We address this concern by building upon prior work that has validated the use of VLMs for this specific task~\cite{kupssinsku2025decodi}, in which it was demonstrated a high statistical agreement between VLM annotations and those from human evaluators, providing confidence in the reliability of this scalable approach for large-scale benchmarking.

\subsubsection{Generalizability and Extensibility}
While this study focuses on the trade-off between text-image alignment and demographic fairness, the framework itself is designed for broad applicability. 
The underlying Pareto-based methodology is metric-agnostic. 
To further enhance reproducibility and efficiency, our open-source tool allows researchers not only to process images from scratch but also to upload previously computed results in a standardized JSON format. 
This decouples the computationally-expensive evaluation step from the interactive analysis, enabling users to dynamically select any two metrics from the saved results and instantly visualize their trade-off landscape. 
This transforms our contribution from a single experiment into a generalizable platform for fostering more rigorous and transparent evaluations across the field of generative AI, discouraging parameter cherry-picking and promoting a deeper understanding of model behavior.

\subsection{Experimental Setup}

Our experimental setup was designed to ensure a rigorous and reproducible comparison across different models and techniques. 
We focused on a benchmark task and systematically explored the hyperparameter space of each model to generate a comprehensive set of data points for the Pareto frontier analysis.

\subsubsection{Models and Benchmark Task}

Our analysis focuses on comparing two debiasing methods with their respective baseline models:

\begin{itemize}
    \item \textbf{Stable Diffusion 1.5 (SD1.5)} as a baseline, compared against \textbf{Fair Diffusion}.
    \item \textbf{Stable Diffusion XL 1.0 (SDXL1.0)} as a baseline, compared against \textbf{DeCoDi}.
    \item \textbf{FLUX-dev}, evaluating different configurations against its default hyperparameters
    \item Comparing all models.
\end{itemize}

Across all experiments, we used the prompt ``\texttt{The face of a nurse}''. This prompt is widely used in the literature because it is known to reveal significant occupational gender and racial biases in T2I models, making it an ideal test case for our framework.

\subsubsection{Image Generation and Parameter Space}

To construct the trade-off landscapes for each model, we purposefully varied key hyperparameters known to influence the generation process. This approach was designed to generate a diverse set of configurations spanning the fairness-utility spectrum.

For each debiasing method (Fair Diffusion, DeCoDi) and for the FLUX model, we generated $5,000$ images. 
These were organized into $50$ distinct hyperparameter configurations, with $100$ images generated per configuration with $100$ different seeds. 
The specific hyperparameters varied were:

\begin{itemize}
    \item \textbf{For Fair Diffusion}, we used the \textit{editing\_concept} of \texttt{["male person", "female person"]} to encourage gender diversity. 
    We varied the \textit{guidance\_scale} across values of $[0.0, 7.0, 10.0, 12.0, 15.0]$ and the \textit{edit\_guidance\_scale} across $[0, 1, 2, 3, 5, 7, 9, 11, 13, 15]$.

    \item \textbf{For DeCoDi}, we used the \textit{safety\_text\_concept} for ``nurse'' as presented in the original work \cite{kupssinsku2025decodi}. 
    We created combinations by varying \texttt{guidance\_scale} across $[0.0, 7.0, 10.0, 12.0, 15.0]$ and \textit{sld\_threshold} across $[0.0, 0.005, 0.01, 0.015, 0.02, 0.025, 0.03, 0.04, 0.05, 1.0]$.

    \item \textbf{For FLUX-dev}, we varied the classifier-free guidance \textit{cfg} parameter across $[1.0, 2.0, 3.5, 4.0, 5.0, 6.0, 7.0, 8.0]$ and the \textit{num\_steps} across $[20, 40, 50, 60, 70]$.
\end{itemize}

The hyperparameters that are varied for these models were specifically chosen based on their strong, documented influence on the generation process, as evidenced by prior work~\cite{greenberg2025demystifying, kupssinsku2025decodi, friedrich2024auditing}. 
This deliberate selection focuses on parameters known to directly impact the fairness-utility trade-off, a central goal of our experiments.

To ground the experimental analysis, we established reference points using the baseline models. 
For \textit{SD 1.5}, \textit{SDXL 1.0}, and \textit{FLUX-dev (default)}, we generated a single datapoint consisting of $1,000$ images each. 
These were produced using the models standard, out-of-the-box parameter settings, providing a clear benchmark against which the performance of the debiasing techniques and their various configurations can be measured.

\subsubsection{Evaluation Pipeline}

The evaluation pipeline, implemented as a Streamlit application, is central to the proposed framework. 
It offers two primary workflows: users can either upload folders of images for on-the-fly processing or load pre-computed results from a standardized JSON file. 
This second option is particularly useful for large-scale experiments, as it decouples the computationally intensive evaluation from the interactive analysis. 
We provide the script for the robust batch processing that produces these JSON files.

Once the data is loaded, the application visualizes the results. 
All $50$ configurations for a given model are plotted on a 2D scatter plot, which we term \textit{Evaluated Configurations}. 
The tool then computes and highlights the \textit{Pareto-optimal frontier} for these points. 
Users can select any two metrics from the available data to serve as the axes for the trade-off analysis.

The tool also allows for overlaying multiple datasets. 
For instance, the Pareto frontier of Fair Diffusion can be plotted alongside the single-point configuration of its baseline, SD~1.5. This feature enables direct, quantitative comparisons, making it clear whether a new method offers a superior trade-off curve or whether its performance is dominated by the baseline. 
To enhance usability, the application displays a table of the Pareto-optimal configurations with their specific hyperparameter values and allows results to be exported in both JSON and CSV formats. 
For direct image processing, the application requires an API key for a VLM judge to perform the demographic annotation.


\section{Results and Discussion}

We demonstrate the proposed framework through experiments that map the fairness-utility trade-offs of prominent T2I models and debiasing methods, presenting the protocol and key findings below.

\subsection{Benchmarking Fairness-utility Trade-offs}
Leveraging the multi-objective optimization framework described in the Methodology section, we now present the empirical evaluation of fairness-utility trade-offs across T2I models and debiasing techniques in three distinct case studies.

\subsubsection{Case Study 1: Fair Diffusion vs SD 1.5}
\begin{figure*}[t]
    \centering 

    \begin{subfigure}[b]{\columnwidth}
        \centering
        \includegraphics[width=\linewidth]{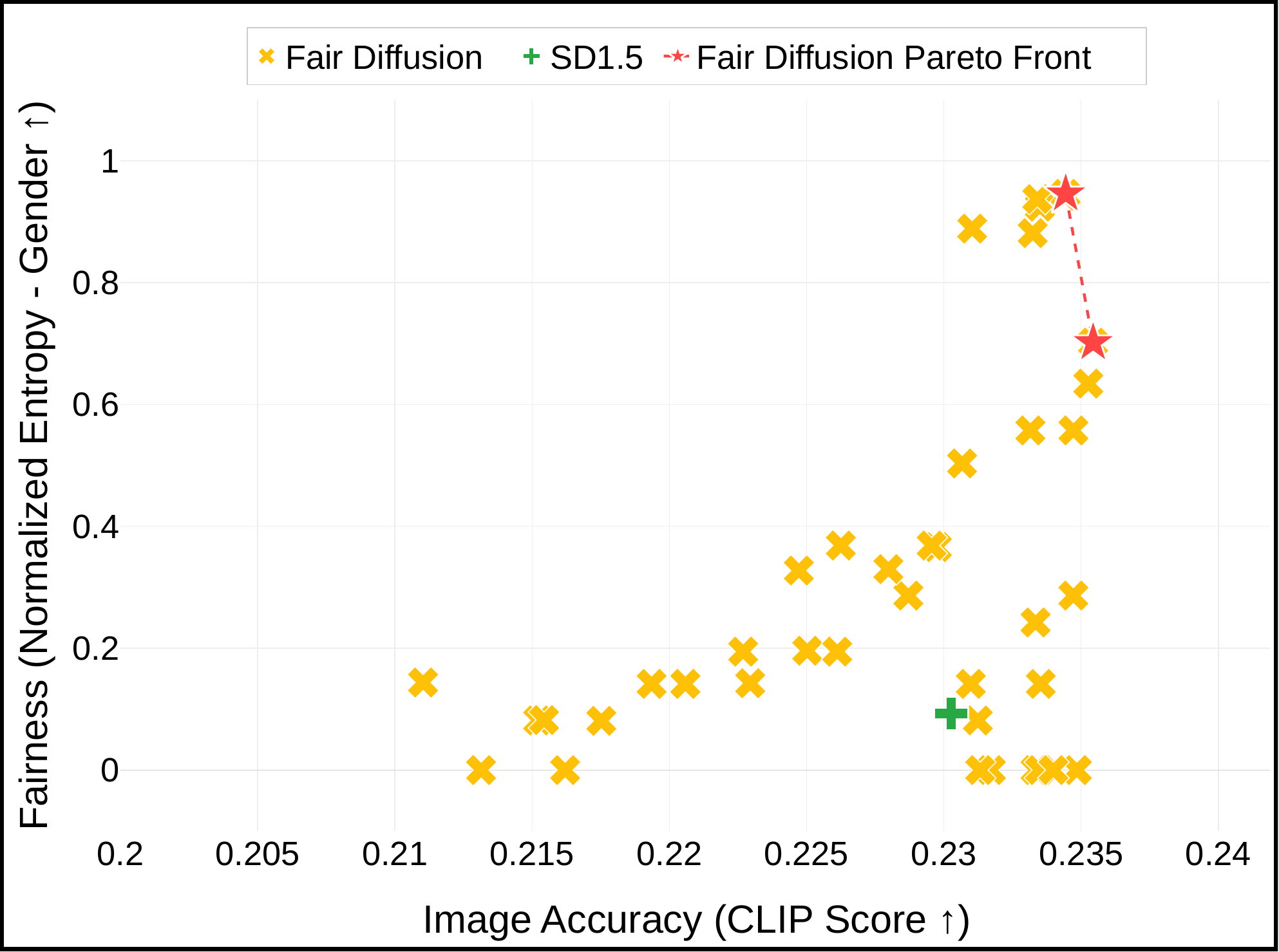}
        \caption{Fair Diffusion vs. SD 1.5}
        \label{fig_fair_sd}
    \end{subfigure}
    \hfill 
    \begin{subfigure}[b]{\columnwidth}
        \centering
        \includegraphics[width=\linewidth]{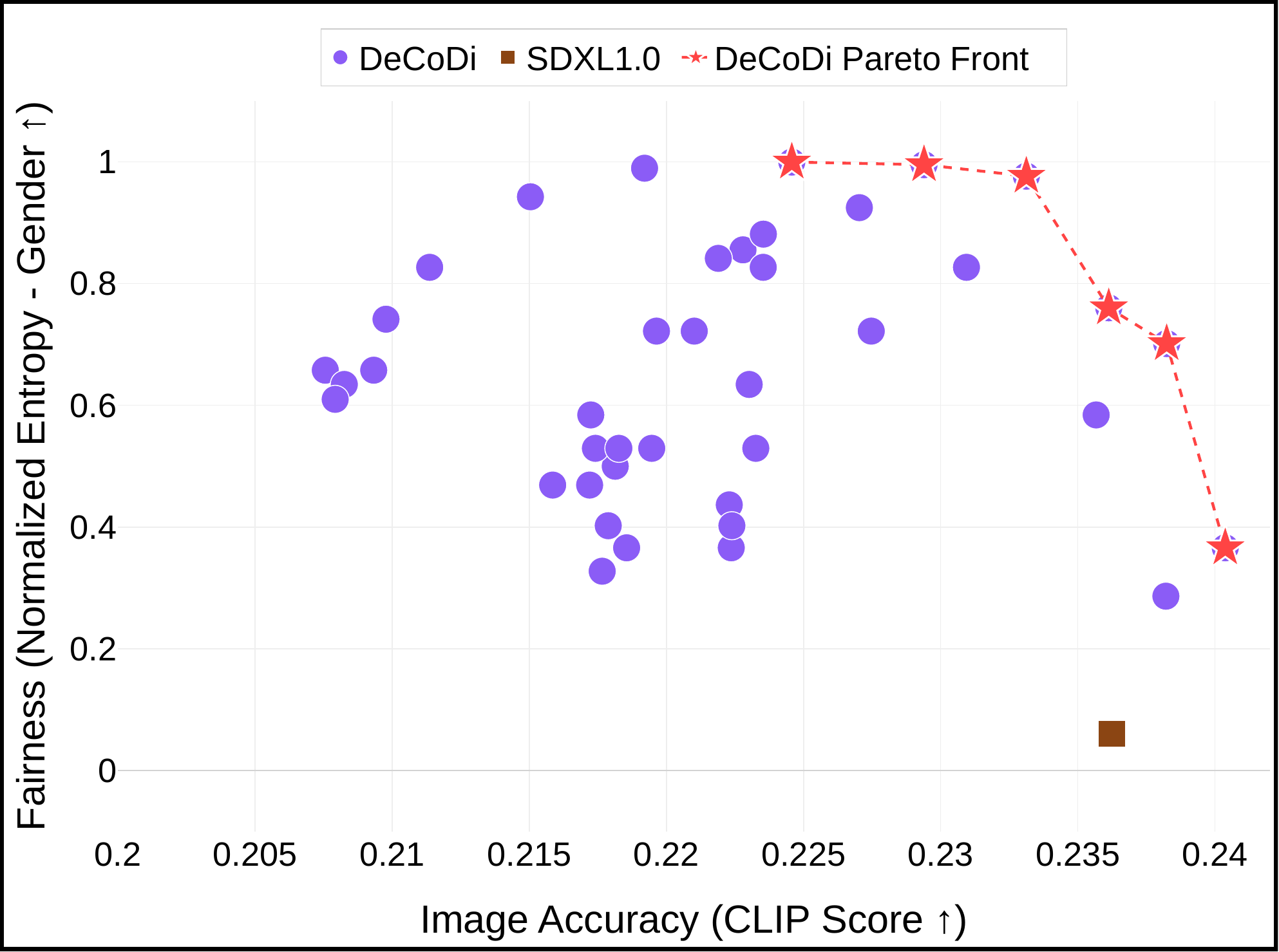}
        \caption{DeCoDi vs. SDXL 1.0}
        \label{fig_decodi_sdxl}
    \end{subfigure}

    \vspace{1em} 

    \begin{subfigure}[b]{\columnwidth}
        \centering
        \includegraphics[width=\linewidth]{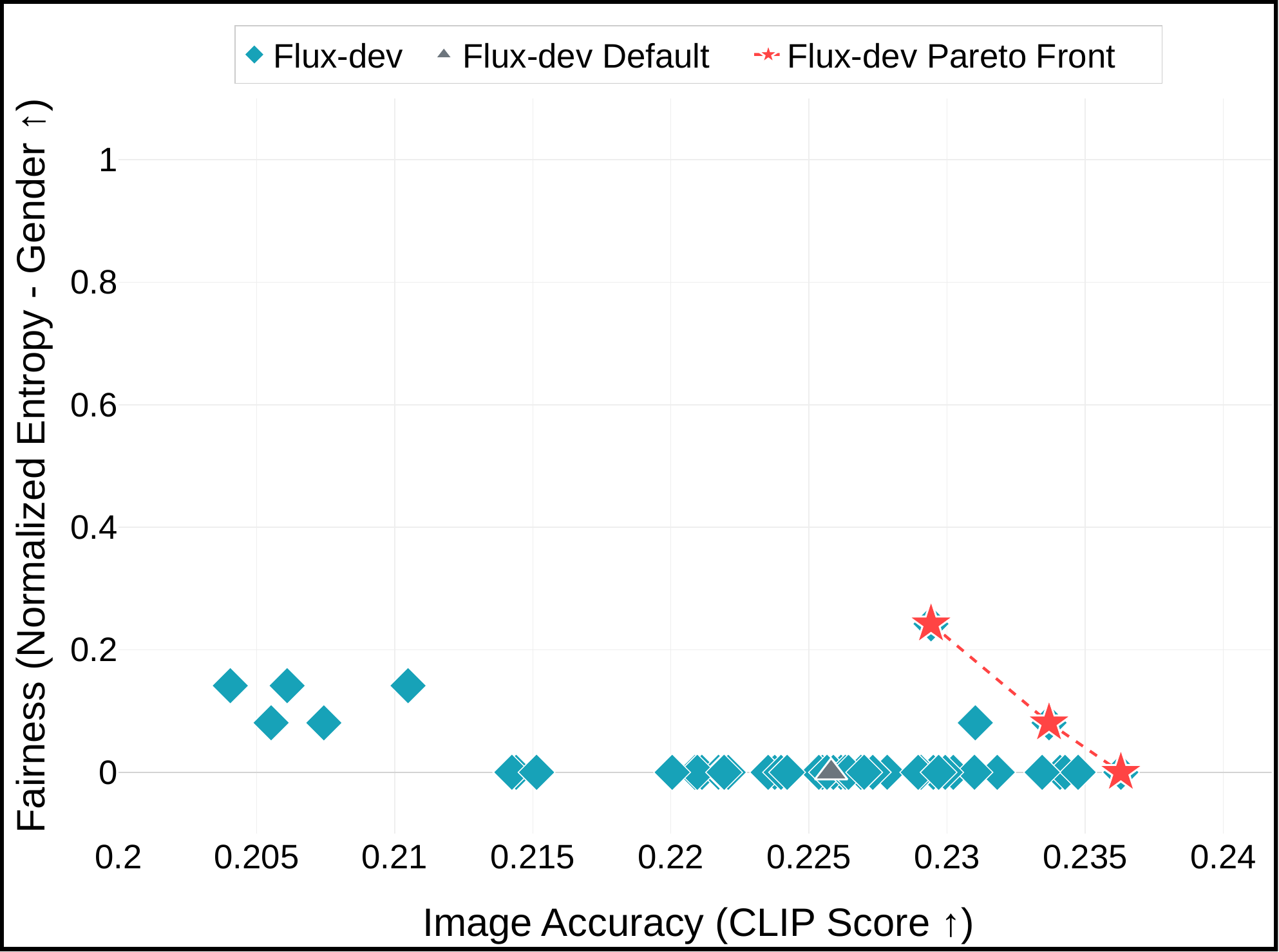}
        \caption{FLUX-dev configurations}
        \label{fig_flux_configs}
    \end{subfigure}
    \hfill
    \begin{subfigure}[b]{\columnwidth}
        \centering
        \includegraphics[width=\linewidth]{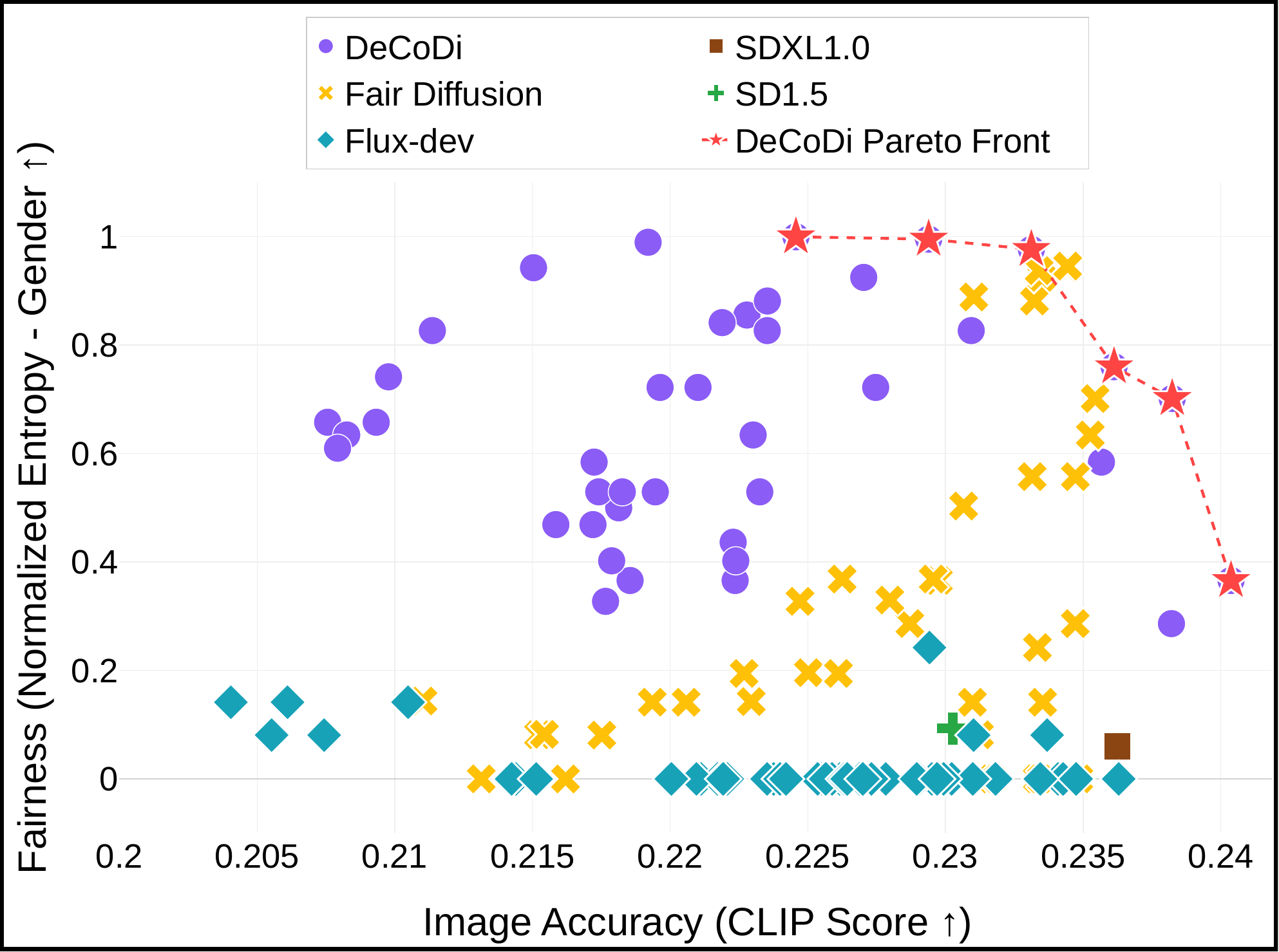} 
        \caption{DeCoDi's Pareto Frontier against other debiasing methods}
        \label{fig_decodi_vs_all}
    \end{subfigure}

    \caption{Fairness-utility Pareto frontiers for base models in comparison to debiased counterparts. (a) Fair Diffusion vs. its SD 1.5 baseline. (b) DeCoDi vs. its SDXL 1.0 baseline. (c) FLUX-dev vs. its own default hyperparameters. (d) Overall comparison.}
    \label{fig_combined_frontiers} 
\end{figure*}

Fair Diffusion~\cite{friedrich2024auditing} operates by applying semantic guidance to mitigate biases during the generation process. 
To explore its fairness-utility trade-offs, we varied two hyperparameters: the Classifier-Free Guidance (CFG) scale and the fair guidance parameter ($\gamma$), seeking the optimal configurations that maximize both metrics.
Using the proposed framework, we generated $50$ distinct data points. As illustrated in Figure~\ref{fig_fair_sd}, the analysis for gender diversity revealed a Pareto frontier comprising just two configurations. This indicates that these two settings dominate the other $48$ variations in terms of their combined fairness and utility performance.

We benchmarked Fair Diffusion, which is based on the Stable Diffusion 1.5 (SD1.5) architecture, against the baseline model. The visualization makes it clear that the dominant configurations from Fair Diffusion surpass the SD1.5 reference point on both fairness ($y$-axis) and utility ($x$-axis). 
This result demonstrates that with proper tuning, guided by the framework, it is possible to find configurations that strictly dominate the baseline model, improving both fairness and utility simultaneously. 

In this case, there is no objective reason to opt for SD1.5 regarding fairness or utility criteria, since Fair Diffusion offers solutions that dominate its baseline counterpart model.

\subsubsection{Case Study 2: DeCoDi vs SDXL 1.0}

In our second case study, we apply the proposed framework to compare the debiasing method DeCoDi against its baseline model, SDXL 1.0. 
We explored DeCoDi's performance landscape by generating multiple configurations, varying its guidance scale and its safety threshold parameter ($\lambda$), which controls the strength of the debiasing.

Results are presented in the trade-off visualization in Figure~\ref{fig_decodi_sdxl}. 
The analysis identified a Pareto frontier for DeCoDi consisting of six distinct hyperparameter configurations. 
This frontier clearly illustrates the trade-off between maximizing fairness and utility. 
For instance, one optimal configuration achieves an almost perfect fairness score (Normalized Entropy for gender $\approx 1.0$), while the configuration with the highest image utility (CLIP Score $\approx 0.24$) yields a lower fairness score of approximately $0.37$.

The comparison shows that the default SDXL~1.0 configuration is dominated by the entire DeCoDi Pareto front. 
While the baseline SDXL~1.0 model achieves a relatively high utility score, its fairness score is smaller than $0.1$. 
The framework reveals multiple DeCoDi settings that offer both slightly higher utility and vastly superior fairness. 
This analysis confirms that DeCoDi provides a significantly improved set of configurations, allowing users to select one that offers a substantially better and more principled balance between fairness and utility than the standard model.

\subsubsection{Case Study 3: FLUX-dev bias}

The final case study provides an intrinsic analysis of the Flux-dev architecture, mapping its inherent fairness-utility landscape. 
We varied its \textit{cfg} and \textit{n\_steps} hyperparameters, using the model's own default configuration as a baseline for comparison.

Results are presented in Figure~\ref{fig_flux_configs}. 
They reveal a strong inherent bias. 
While FLUX-dev achieves competitive utility (CLIP Scores up to $0.24$), its fairness scores remain extremely low, with a maximum Normalized Entropy of only $0.25$. 
It is reflected in a limited and steep Pareto frontier, indicating a poor trade-off where minor fairness gains require a significant sacrifice in utility. 
This case study thus demonstrates that architectural advancements alone do not eliminate social biases, proving that dedicated debiasing techniques remain crucial.

\subsection{Discussion and Key Insights}

To evaluate the performance of different models in jointly optimizing fairness (measured by Entropy) and utility (measured by CLIP Score), we generated Pareto frontiers for each model in our experiments. 
Subsequently, we identified the hyperparameter configurations that yielded the points belonging to the Pareto frontier of each model. 
These points represent the best trade-offs between fairness and utility achieved by each architecture. 
Table~\ref{tab:model_performance} presents the hyperparameter configurations and their respective CLIP Score and Entropy values for the points on each model's Pareto frontier.

\begin{table}[b]
\caption{Pareto Frontier configurations for the case studies.}
\label{tab:model_performance}
\centering
\resizebox{\columnwidth}{!}{
\begin{tabular}{lcccl}
\toprule
\textbf{Model} & \textbf{Config.} & \textbf{ClipScore} & \textbf{Entropy} & \textbf{Hyperparameters} \\
\midrule
DeCoDi & $1$ & $0.24$ & $0.366$ & cfg $=15.0$, $\lambda = 0.0$ \\
DeCoDi & $2$ & $0.229$ & $0.995$ & cfg $=12.0$, $\lambda = 0.01$ \\
DeCoDi & $3$ & $0.236$ & $0.76$ & cfg$=12.0$, $\lambda =0.005$ \\
DeCoDi & $4$ & $0.238$ & $0.70$1 & cfg$=15.0$, $\lambda =0.005$ \\
DeCoDi & $5$ & $0.233$ & $0.977$ & cfg$=15.0$, $\lambda =0.01$ \\
DeCoDi & $6$ & $0.225$ & $0.999$ & cfg$=10.0$, $\lambda =0.01$ \\
Fair Diffusion & $1$ & $0.235$ & $0.701$ & cfg$=12.0$, $\gamma =3.0$ \\
Fair Diffusion & $2$ & $0.234$ & $0.946$ & cfg$=12.0$, $\gamma =5.0$ \\
FLUX-dev & $1$ & $0.234$ & $0.081$ & cfg$=8.0$, n\_steps$=20$ \\
FLUX-dev & $2$ & $0.229$ & $0.242$ & cfg$=7.0$, n\_steps$=20$ \\
FLUX-dev & $3$ & $0.236$ & $0$ & cfg$=2.0$, n\_steps$=20$ \\
FLUX-dev default & $1$ & $0.225$ & $0$ & default \\
SDXL1.0 & $1$ & $0.236$ & $0.06$ & default \\
SD1.5 & $1$ & $0.23$ & $0.09$ & default \\
\bottomrule
\end{tabular}%
}
\end{table}

\subsubsection{Metric Dominance vs. Perceptual Quality}

A key strength of our framework is its ability to holistically compare multiple models on the same trade-off landscape. 
To demonstrate this fact, we aggregated the configurations from all evaluated models, including the configurations from all evaluated models: debiasing methods (Fair Diffusion, DeCoDi), the state-of-the-art FLUX-dev, and the baseline points (SD1.5, SDXL1.0) into a single visualization, presented in Figure~\ref{fig_decodi_vs_all}.


A purely metric-driven analysis reveals that the Pareto-optimal frontier of DeCoDi almost entirely dominates the configurations of all other models. 
This suggests that DeCoDi offers the most efficient fairness-utility trade-off, achieving higher fairness for any given level of utility.

However, a quantitative analysis alone can be misleading. 
A manual, qualitative inspection of the generated images (see Figure~\ref{fig:intro_imgs_single_point}), reveals a critical limitation. 
While DeCoDi excels in the metrics, its generated images often suffer from poor perceptual quality, appearing distorted or ``cartoonish''. 
This suggests that metrics like CLIPScore, while effective at capturing semantic alignment, may not fully penalize certain types of distortions that are immediately obvious to a human observer. 
In contrast, models like Flux-dev, while dominated on the 2D metric plot, produce images of significantly higher visual fidelity. 
Consequently, relying solely on a 2D Pareto front can lead to the selection of models that are metrically optimal but practically unusable.

This discrepancy between quantitative results and visual artifacts points to a clear direction for future work. 
We propose extending our framework to a third dimension with a metric for perceptual quality, such as the Fréchet Inception Distance (FID). 
This would require evolving our visualizations from 2D plots to 3D representations to compare all three objectives.
Addressing this is a valuable research opportunity to build more robust tools for the responsible evaluation of fairness and utility in generative models.

\section{Conclusion}
This work proposed a benchmark framework that uses Pareto-optimal frontiers to address the limitations of single-point evaluations in Text-to-Image (T2I) models, treating the fairness-utility trade-off as a multi-objective optimization problem. 
Applying the proposed approach to baseline models, debiasing techniques, and state-of-the-art architectures, we demonstrated that default configurations are consistently dominated by alternatives that offer a superior fairness-utility trade-off. 
Our analysis also reveals that metric-based dominance does not always align with perceptual quality, highlighting the need to extend such evaluations to a third dimension, like a perceptual fidelity score, for a more complete assessment. 
Ultimately, our framework provides a more rigorous and transparent method for model selection, moving the field toward a more responsible evaluation of generative AI.

\bibliography{aaai25}

\end{document}